\documentclass[10pt,twocolumn,letterpaper]{article}

\usepackage{3dv}
\usepackage{times}
\usepackage{epsfig}
\usepackage{graphicx,wrapfig,lipsum}
\usepackage{graphicx}
\usepackage{amsmath}
\usepackage{bbm}
\usepackage{amssymb}
\usepackage{booktabs} 
\usepackage{array} 
\usepackage{paralist} 
\usepackage{authblk}
\usepackage{verbatim} 
\usepackage{subfig} 
\usepackage{diagbox}
\usepackage{multirow}
\newcommand{\norm}[1]{\left\lVert#1\right\rVert}

\usepackage[pagebackref=true,breaklinks=true,letterpaper=true,colorlinks,bookmarks=false]{hyperref}

\threedvfinalcopy 

\def\threedvPaperID{129} 

\ifthreedvfinal\fi
\begin{document}
\title{DeepHPS: End-to-end Estimation of 3D Hand Pose and Shape by Learning from Synthetic Depth\vspace{-4ex}}

\author[1,3]{Jameel Malik}
\author[1]{Ahmed Elhayek}
\author[2]{Fabrizio Nunnari}
\author[1]{Kiran Varanasi}
\author[2]{\\Kiarash Tamaddon}
\author[2]{Alexis Heloir}
\author[1]{Didier Stricker\vspace{-1ex}}
\affil[1]{\it AV group, DFKI Kaiserslautern, Germany}\affil[2]{\it DFKI-MMCI, SLSI group, Saarbruecken, Germany}\affil[3]{\it NUST-SEECS, Pakistan\vspace{-5ex}}

\clearpage\maketitle
\thispagestyle{empty}


\begin{abstract}
\vspace{-2ex}
Articulated hand pose and shape estimation is an important problem for vision-based applications such as augmented reality and animation. 
 In contrast to the existing methods which optimize only for joint positions, we propose a fully supervised deep network which learns to jointly estimate a full 3D hand mesh representation and pose from a single depth image. To this end, a CNN architecture is employed to estimate parametric representations i.e. hand pose, bone scales and complex shape parameters. Then, a novel hand pose and shape layer, embedded inside our deep framework, produces 3D joint positions and hand mesh. 
Lack of sufficient training data with varying hand shapes limits the generalized performance of learning based methods. Also, manually annotating real data is suboptimal. 
Therefore, we present SynHand5M: a million-scale synthetic dataset with accurate joint annotations, segmentation masks and mesh files of depth maps.
Among model based learning (hybrid) methods, we show improved results on our dataset and two of the public benchmarks i.e. NYU and ICVL. Also, by employing a joint training strategy with real and synthetic data, we recover 3D hand mesh and pose from real images in 3.7ms.    
\end{abstract}

\vspace{-6mm}
\section{Introduction}
\vspace{-1mm}
3D hand pose estimation is essential for many computer vision applications such as activity recognition, human-computer interaction and modeling user intent. However, the advent of virtual and augmented reality technologies makes it necessary to reconstruct the 3D hand surface together with the pose. Recent years have seen a great progress in the pose estimation task primarily due to significant developments in deep learning and the availability of low cost commodity depth sensors. 
However, the stated problem is still far from being solved due to many challenging factors that include large variations in hand shapes, view point changes, many degrees of freedom (DoFs), constrained parameter space, self similarity and occlusions.

\begin{figure}
\begin{center}
    \setlength\fboxsep{0.6pt}
    \setlength\fboxrule{0.6pt}
      \includegraphics[width=0.95\linewidth]{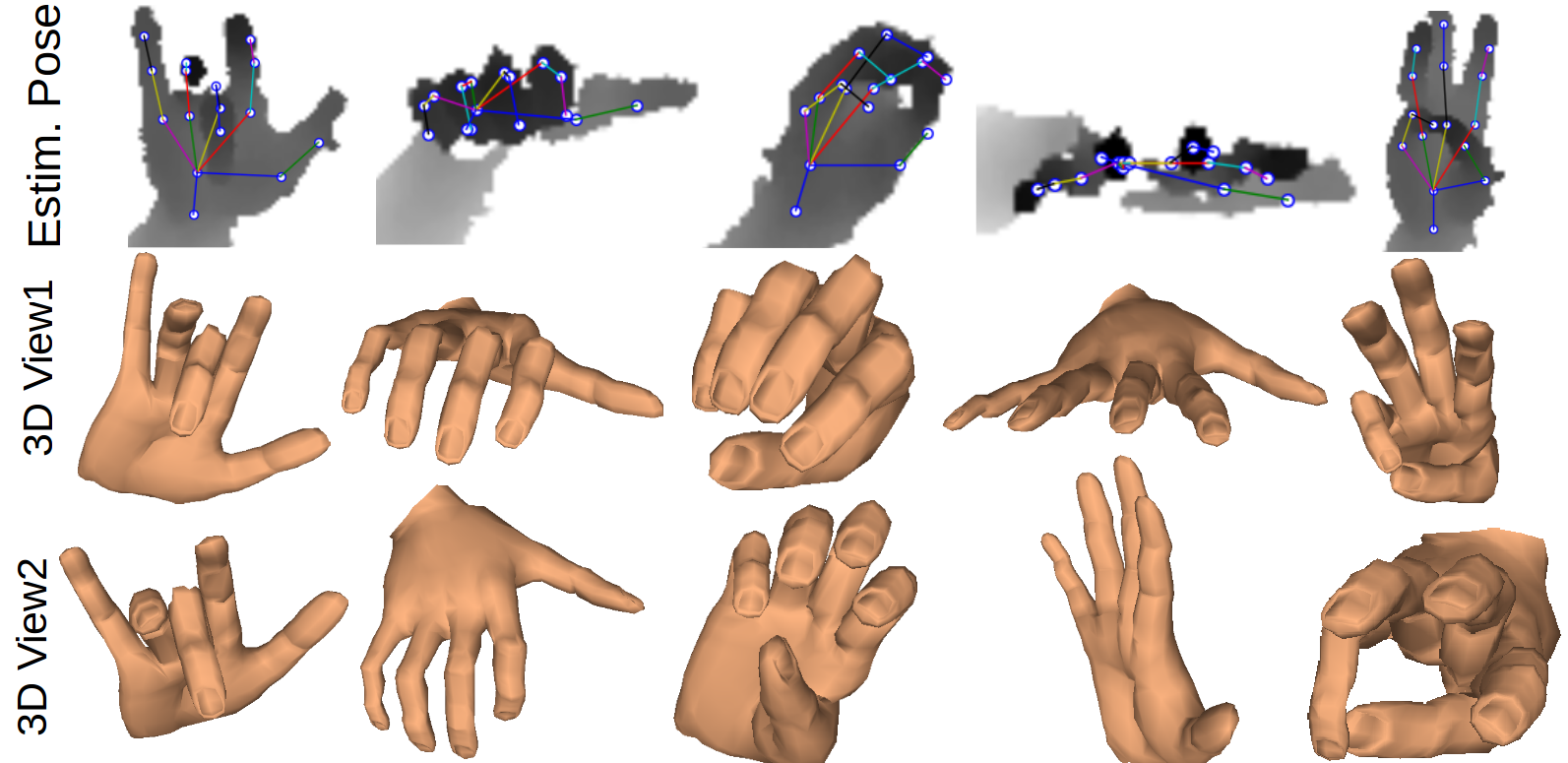}
\end{center}
    \vspace{-5mm}
   \caption{ {\bf{Real hand pose and shape recovery}}: We describe a deep network for recovering the 3D hand pose and shape of NYU\cite{tompson2014real} depth images by learning from synthetic depth. Note that we infer 3D pose and shape even in cases of missing depth and occluded fingers.\vspace{-6mm}}
  
\label{fig:figure_title}
\end{figure} 

\begin{figure}[t]
\begin{center}
	\setlength\fboxsep{0.6pt}
	\setlength\fboxrule{0.6pt} 

    \includegraphics[width=1.3cm,height=1.48cm]{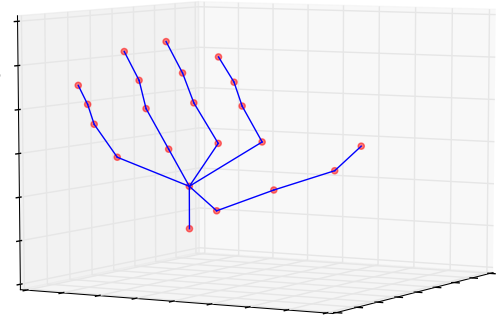}\;
 	\includegraphics[width=1.3cm,height=1.48cm]{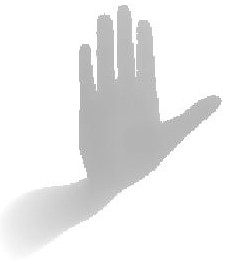}\;\;\;\;\;\;\;\;
    \includegraphics[width=1.2cm,height=1.48cm]{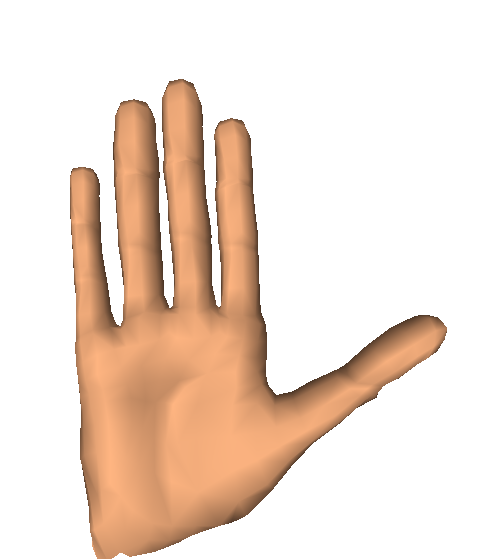}
 	\includegraphics[width=1.2cm,height=1.48cm]{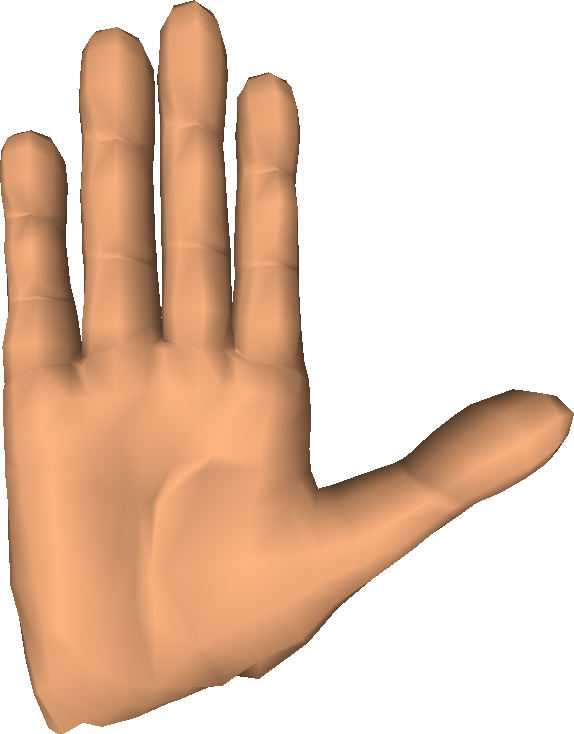}
	\includegraphics[width=1.2cm,height=1.48cm]{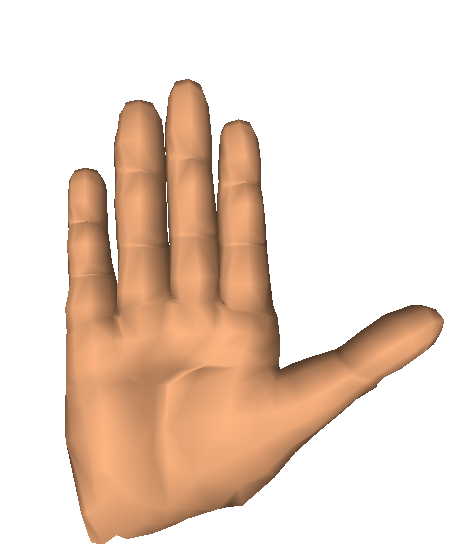}
    \\
    \includegraphics[width=1.3cm,height=1.48cm]{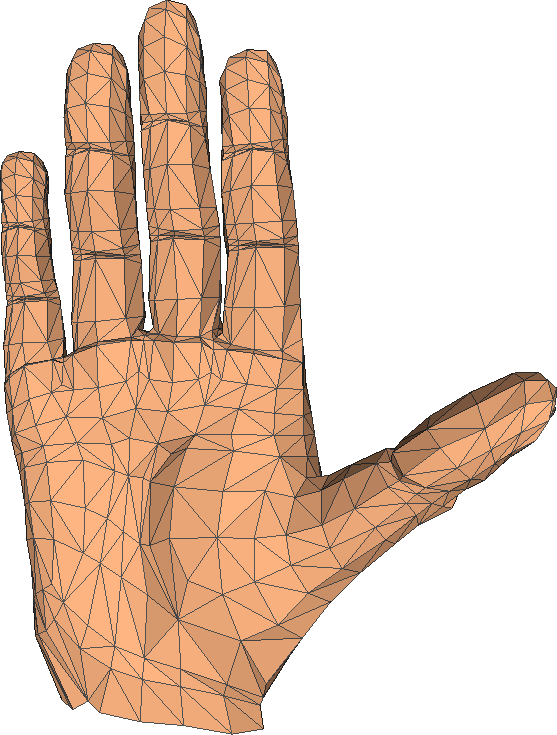}
	\includegraphics[width=1.3cm,height=1.48cm]{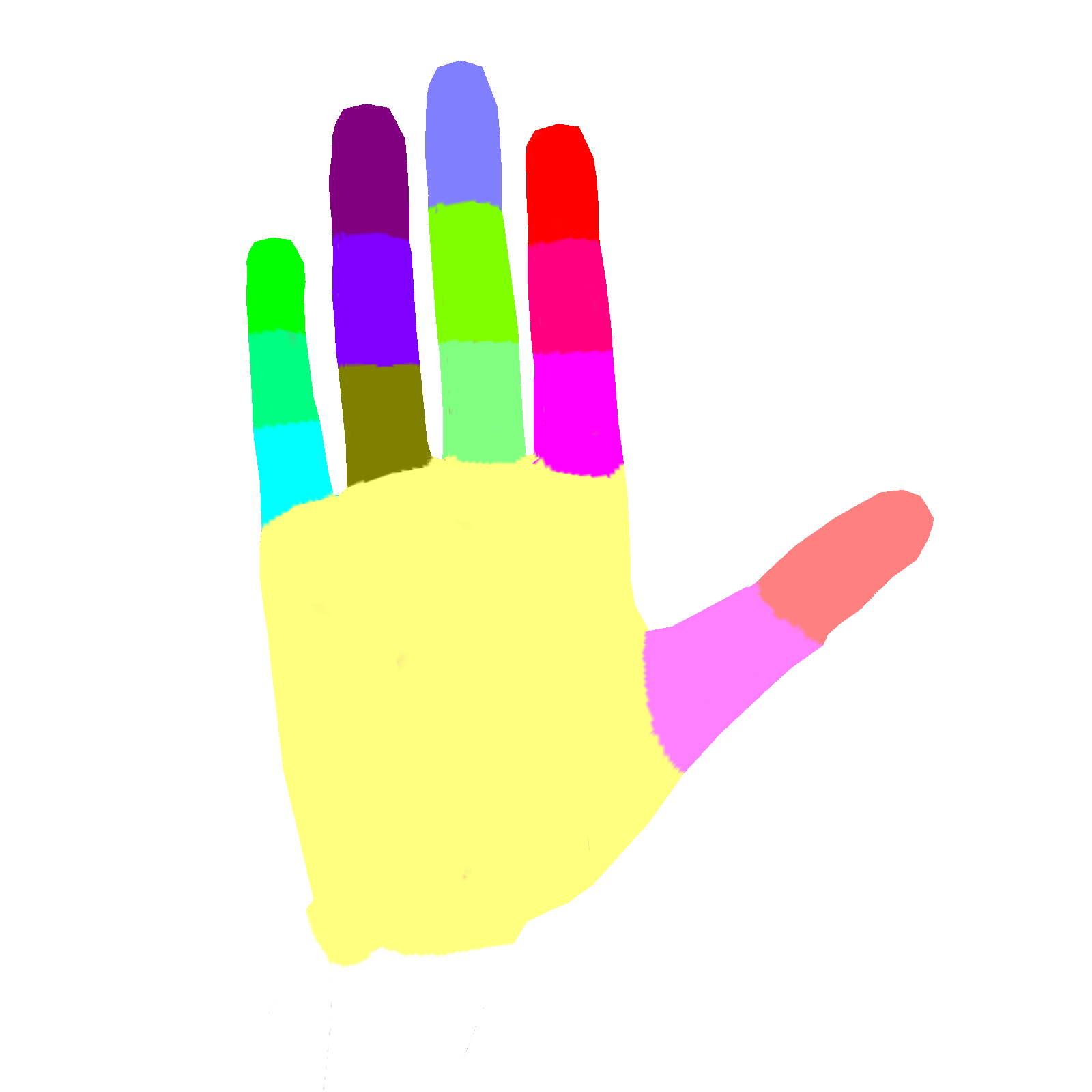}\;\;\;\;\;\;\;\;\;
    \includegraphics[width=1.2cm,height=1.48cm]{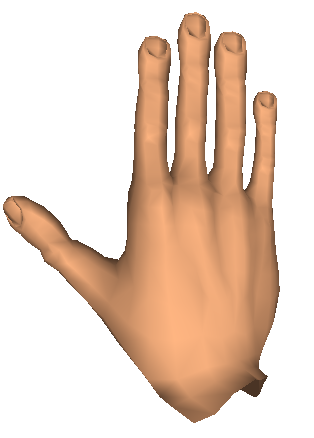} 
    \includegraphics[width=1.2cm,height=1.48cm]{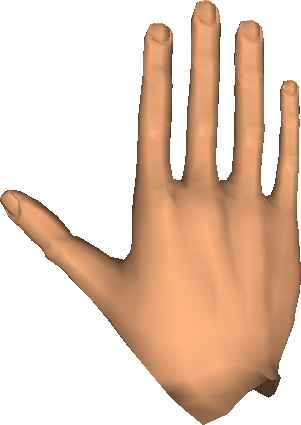}
	\includegraphics[width=1.2cm,height=1.48cm]{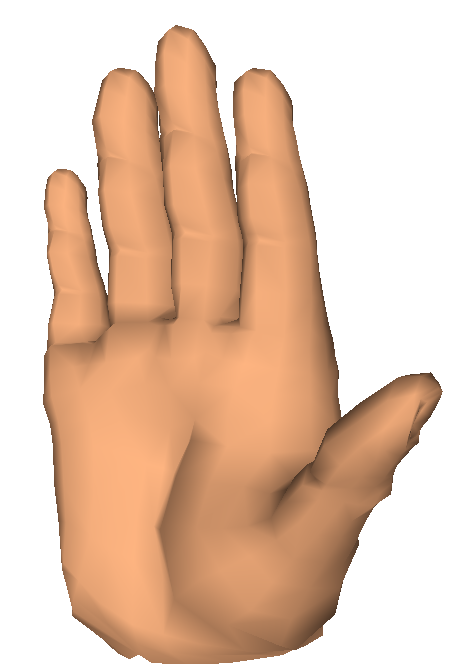}\\
     (a) Dataset components \;\;\;\;\;\;(b) Shape variations \;\;\;\;\;\;
\end{center}
\vspace{-5mm}
   \caption{The SynHand5M dataset contains $5$ million images. (a) The dataset ground truth components: hand poses (joints angles and 3D positions), depth maps, mesh files, and hand parts segmentation. (b) Samples illustrating the big variation in shape.
    }
\label{fig:datasetVC}
\vspace{-5mm}
\end{figure}

Large amounts of training data, enriched with all possible variations in each of the challenging aspects stated above, are a key requirement for deep learning based methods to generalize well and achieve significant gains in accuracy. 
The recent real dataset \cite{yuan2017bighand2} gathers a sufficient number of annotated images. However, it is very limited in hand shape variation (i.e. only 10 subjects). 
Progress in essential tasks such as estimation of hand surface and hand-part segmentation is hampered, as manual supervision for such problems at large scale is extremely expensive. 
In this paper, we generate a synthetic dataset that addresses these problems. It not only allows us to create virtually infinite training data, with large variations in shapes and view-points, but it also produces annotations that are highly accurate even in the case of occlusions. One weakness of synthetic datasets is their limited realism. A solution to this problem has been proposed by \cite{shrivastava2017learning, mueller2017ganerated}, where a generative adversarial training network is employed to improve the realism of synthetic images. However, producing realistic images is not the same problem as improving the recognition rates of a convolutional neural network (CNN) model. In this paper, we address this latter problem, and specifically focus on a wide variation of hand shapes, including extreme shapes that are not very common (in contrast to ~\cite{MANO:SIGGRAPHASIA:2017}).
We present SynHand5M: a new million scale synthetic dataset with accurate ground truth joints positions, angles,  mesh files, and segmentation masks of depth frames; see Figure  \ref{fig:datasetVC}. 
Our SynHand5M dataset opens up new possibilities for advanced hand analysis.

Currently, CNN-based discriminative methods are the state-of-the-art which estimate 3D joint positions directly from depth images \cite{guo2016two,oberweger2017deepprior++,deng2017hand3d,rad2017feature}. 
However, major weakness of these methods is that the predictions are coarse with no explicit consideration to kinematics and  geometric constraints. Sinha et al. \cite{sinha2017surfnet} propose to estimate 3D shape surface from depth image or hand joint angles, using a CNN. However, their approach neither estimates hand pose nor considers kinematics and physical constraints.  
Also, these methods generalize poorly to unseen hand shapes \cite{yuan2018depth}. 

On the other hand, building a personalized hand model requires a different generative approach, that optimizes a complex energy function to generate the hand pose \cite{Roditakis2017,oikonomidis2011efficient,qian2014realtime,tagliasacchi2015robust,tang2015opening}. However, person specific hand model calibration clearly restricts the generalization of these methods for varying hand shapes.   
Hybrid methods combine the advantages of both discriminative and generative approaches \cite{ge2016robust,sinha2016deephand,oberweger2015training,tang2014latent,sridhar2016real,ye2017occlusion,zhang2016learning}. 
To the best of our knowledge, none of the existing works explicitly addresses the problem of jointly estimating full hand shape surface, bone-lengths and pose in a single deep framework. 

In this paper, we address the  problem of generalizing 3D hand pose and surface geometry estimation over varying hand shapes. We propose to embed a novel hand pose and shape layer (HPSL) inside deep learning network to jointly optimize for 3D hand pose and shape surface. The proposed CNN architecture simultaneously estimates the hand pose parameters, bones scales and shape parameters. All these parameters are fed to the HPSL which implements not only a new forward kinematics function, but also the fitting of a morphable hand model and linear blend skinning to produce both 3D joint positions and 3D hand surface; see Figure \ref{fig:application}. The whole pipeline is trained in an end-to-end manner. In sum, our contributions are:
\begin{enumerate}
\item  A novel deep network layer which performs: 
  \begin{enumerate}
    \item Forward kinematics using a new combination of hand pose and bone scales parameters.
    \item Reconstruction of a morphable hand model from hand shape parameters and the morph targets.
    \item Linear blend Skinning algorithm to animate the 3D hand surface; see Section \ref{ssec:HPSL}.   
  \end{enumerate}
\item  A novel end-to-end framework for simultaneous hand pose and shape estimation; see Section \ref{sec:Overview}.
\item A new $5$ million scale synthetic hand pose dataset that offers accurate ground truth joint angles, 3D joint positions, 3D mesh vertices, segmentation masks; see Section \ref{sec:SYN}. The synthetic dataset will be publicly available.

\end{enumerate}

\section{Related Work}
\label{sec:related}
Depth-based hand pose estimation has been extensively studied in the computer vision community. We refer the reader to the survey \cite{supancic2015depth} for a detailed overview of the field.
Recently, a comprehensive analysis and investigation of the state-of-the-art along-with future challenges have been presented by \cite{yuan2018depth}. 
The approaches can be roughly divided into generative, discriminative and hybrid methods.
In this section, we briefly review the existing hand pose benchmarks. Then, we focus our discussion on CNN-based discriminative and hybrid methods.  

\textbf{Existing Benchmarks.}
Common shortcomings in existing real hand datasets are low variation in hand shape, inaccurate ground truth annotations,  insufficient amount of training data, low complexity (e.g. occlusion) of hand poses, and limited view point coverage.  
Most commonly used benchmarks are NYU \cite{tompson2014real}, ICVL \cite{tang2014latent} and MSRA15 \cite{sun2015cascaded}. NYU hand pose dataset uses a model-based direct search method for annotating ground truth which is quite accurate. It covers a good range of complex hand poses. However, their training set has single hand shape. ICVL dataset uses a guided Latent Tree Model (LTM) based search method and mostly contains highly inaccurate ground truth annotations. Moreover, it uses only one hand model \cite{yuan2017bighand2}. MSRA15 employs an iterative optimization method \cite{qian2014realtime} for annotating followed by manual refinement. It uses $17$ hand poses, however, it has large view-point coverage. The major limitation of this dataset is its limited size and low annotation accuracy. Recently, Yuan et al. \cite{yuan2017bighand2} propose a million scale real hand pose dataset, but it has low variation in hand shape(i.e. only 10 subjects).
Some other very small real hand pose datasets such as Dexter-1 \cite{sridhar2013interactive}, ASTAR \cite{xu2016estimate} , MSRA14 \cite{qian2014realtime} are not suited for large-scale training.   
Several works focused on creating synthetic hand pose datasets.
MSRC \cite{sharp2015accurate} is a synthetic benchmark however, it has only one hand model and limited pose space coverage. In \cite{sinha2017surfnet,Mueller2017RealTimeHT}, medium-scale synthetic hand datasets are used to train CNN models, but they are  not publicly available. 
Given the hard problem of collecting and annotating a large-scale real hand pose dataset, we propose the first million scale synthetic benchmark which consists of more than 5 million depth images together with ground truth joints positions, angles,  mesh files, and segmentation masks. 



\textbf{CNN-based Discriminative Methods.}
Recent works such as \cite{moon2017v2v,chen2017pose,wang2018region,guo2017region,ge2017robust,rad2017feature} exceed in accuracy over random decision forest (RDF) based discriminative methods \cite{sharp2015accurate, sun2015cascaded,wan2016hand,xu2017lie,li20153d}. A few manuscripts have used either RGB or RGB-D data to predict 3D joint positions \cite{zimmermann2017learning,panteleris2017using,simon2017hand,mueller2017real}. 
In \cite{ge2017robust}, Ge et al. directly regress 3D joint coordinates using a 3D-CNN. Recently, \cite{moon2017v2v} introduced voxel-to-voxel regression framework which exploits a one-to-one relationship between voxelised input depth and output 3D heatmaps. \cite{guo2017region,wang2018region} introduce a powerful region ensemble strategy which integrates the outputs from multiple regressors on different regions of depth input. Chen et al. \cite{chen2017pose} extended \cite{wang2018region} by an iterative pose guided region ensemble strategy. In \cite{sinha2017surfnet}, a discriminative hand shape estimation is proposed. Although the accuracy of these methods is the state-of-the-art, they impose no explicit geometric and physical constraints on the estimated pose. 
Also, these methods still fail to generalize on unseen hand shapes \cite{yuan2018depth}. 

\textbf{CNN-based Hybrid Methods.}
Tompson et al. \cite{tompson2014real} employed CNN for estimating 2D heatmaps. Thereafter, they apply inverse kinematics for hand pose recovery. In extension to this work, \cite{ge2016robust} utilize 3D-CNN for 2D heatmaps estimation and afterwards regress 3D joint positions. 
Oberweger et al. \cite{oberweger2015training} utilize three CNNs combined in a feedback loop to regress 3D joint positions. The network comprises of an initial pose estimator, a synthesizer and finally a pose update network. Ye et al. \cite{ye2016spatial} present a hybrid framework using hierarchical spatial attention mechanism and hierarchical PSO. Wan et al. \cite{wan2017crossing} implicitly model the dependencies in the hand skeleton by learning a shared latent space. In \cite{zhou2016model}, a forward kinematics layer, with physical constraints and a fixed hand model, is implemented in an end-to-end training framework. Malik et al. \cite{malik2017simultaneous} further extend this work by introducing a flexible hand geometry in the training pipeline. The algorithm simultaneously estimates bone-lengths and hand pose. In \cite{wan2017dense}, a multi-task cascade network is  employed  to predict 2D/3D joint heatmaps along-with 3D joint offsets. Dibra et al. \cite{dibra2017refine} introduce an end-to-end training pipeline to refine the hand pose using an unlabeled dataset. All of the above described methods cast the problem of hand pose estimation to 3D joints regression only. Our argument is that given the inherent 3D surface geometry information in depth inputs, a differentiable hand pose and shape layer can be embedded in the deep learning framework to regress not only the 3D joint positions but also, the full 3D mesh of hand.                


\section{Method Overview}
\label{sec:Overview}
\begin{figure*}[t]
\label{fig:2}
\begin{center}
	\setlength\fboxsep{0.6pt}
	\setlength\fboxrule{0.6pt} 
    \includegraphics[width=0.66\linewidth]{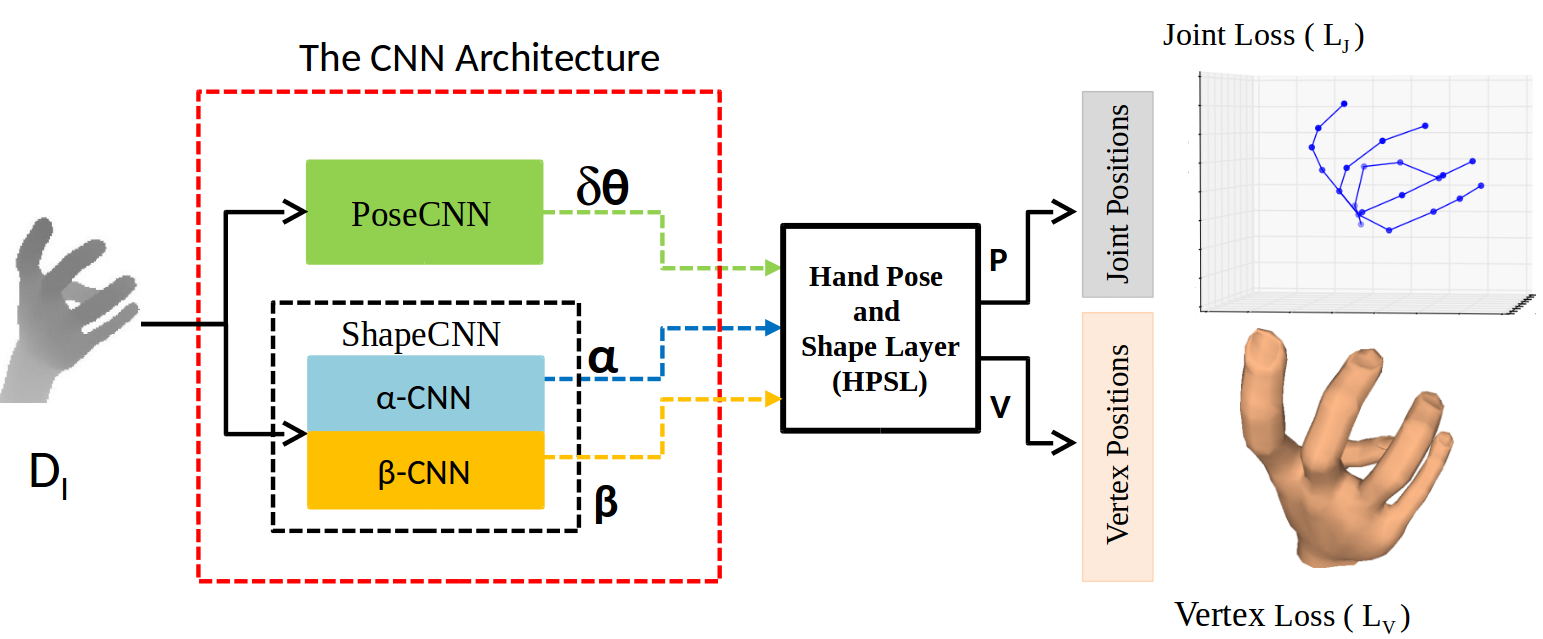} \;\;\;\;\;\;\;\;\;\;\;\;\;
     \includegraphics[width=0.20\linewidth]{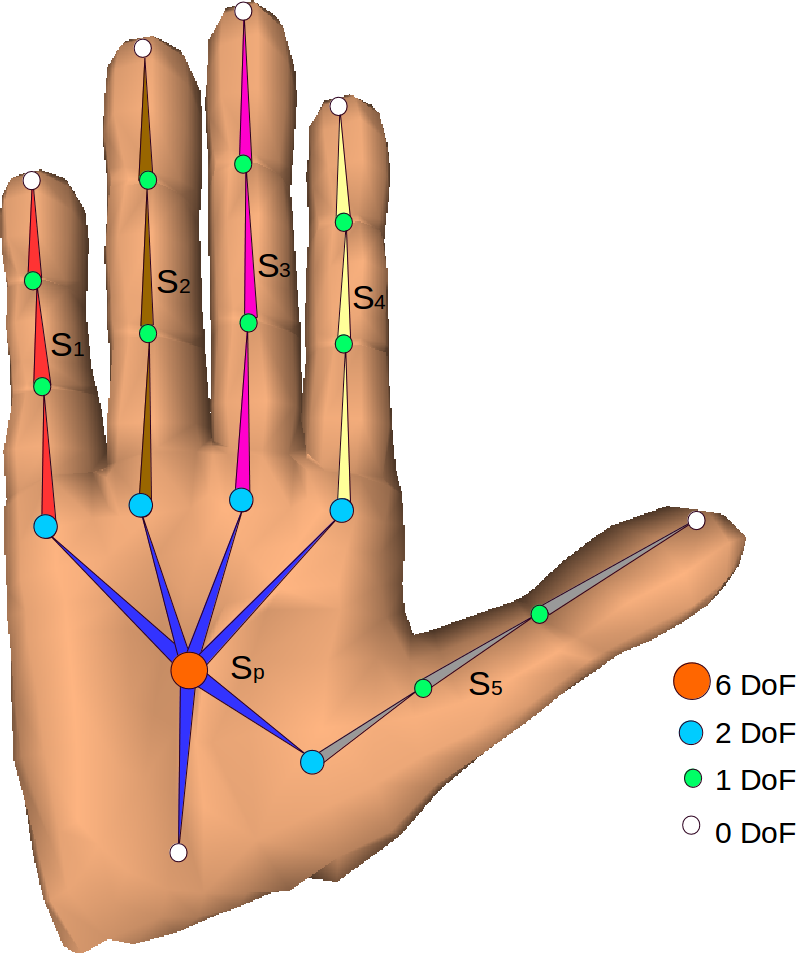}\\
        \hfill
    (a) Algorithm pipeline \;\;\;\;\;\;\;\;\;\;\;\;\;\;\;\;\;\;\;\;\;\;\;\;\;\;\;\;\;\;\;\;\;\;\;\;\;\;\;\;\;\;\;\;\;\;\;\; (b) Our hand model \;\;\;\;\;\;\;\;\;\;\;\;\;
\end{center}
       \vspace{-5mm}
   \caption{ 
   (a) An overview of our method for simultaneous 3D hand pose and surface estimation.
   A depth image $D_I$ is passed through three CNNs to estimate pose parameters $\delta \theta$, bones scales $\alpha$ and shape parameters $\beta$. These parameters are sent to  HPSL which generate the hand joints positions $P$ and hand surface vertices $V$.   
   (b) Our hand model with $26$ DoFs overlaid with the neutral hand shape $b_0$. The bone colors illustrate $6$ bone-length scales $\alpha$.}
\label{fig:application}
\vspace{-5mm}
\end{figure*}

We aim to jointly estimate the locations of $J=22$ 3D hand joints , and $\vartheta=1193$ vertices  of hand mesh from a single depth image $D_I$. Our hand skeleton in rest pose is shown in Figure \ref{fig:application}(b). It has $J$ hand joints defined on $26$ DoFs. The hand root has $6$ DoF; $3$ for global orientation and $3$ for global translation. All other DoFs are defined for joints articulations. The $26$ dimensional pose vector is initialized for the rest pose, called $\theta_{init}$. Any other pose  $\Theta$ can be constructed by adding change $\delta \theta$ to the rest pose i.e. $\Theta$ = $\theta_{init}$ + $\delta \theta$.  
The bone-lengths $B$, are initialized by averaging over all bone-lengths of different hand shapes in our synthetic dataset. In order to add flexibility to the hand skeleton, $6$ different hand bones scales, $\alpha$, are associated to bone-lengths. Our hand mesh has $\vartheta$ vertices and $1184$ faces. The neutral hand surface is shown in Figure \ref{fig:application}(b). We use $7$ hand shape parameters $\beta$ which allow to formulate the surface geometry of a desired hand shape in reference pose; see Section \ref{sec:SYN}.


Our pipeline is shown in Figure \ref{fig:application}(a). Firstly, a new CNN architecture estimates $\delta \theta$, $\alpha$ and $\beta$ given a depth input $D_I$. This architecture consists of PoseCNN which estimates $\delta \theta$ and ShapeCNN which estimates $\alpha$ and $\beta$.
Thereafter, a new non-linear hand pose and shape layer (HPSL) performs forward kinematics, hand shape surface reconstruction and linear blend skinning. The outputs of the layer are   3D joint positions and hand surface vertices. These outputs are used to compute the standard euclidean loss for joint positions and vertices; see Equation \ref{eq:22}. The complete pipeline is trained end-to-end in a fully supervised manner.  



\section{Joint Hand Pose and Shape Estimation}
\label{sec:HPSE}
In this section, we discuss the components of our pipeline which are shown in Figure \ref{fig:application}(a). 
We explain the novel Hand Pose and Shape Layer (HPSL) in detail because it is the main component which allows to jointly estimate hand pose and shape surface.

\subsection{The CNN Architecture}
\label{ssec:HCA}
Our CNN architecture comprises of three parallel CNNs to learn $\delta \theta$, $\alpha$ and $\beta$, given $D_I$. The PoseCNN leverages one of the state-of-the-art CNN \cite{guo2017region} to estimate joint angles $\delta \theta$. However, the CNN was originally used to regress 3D hand joint positions; see Section \ref{sec:related}. We refer the reader to \cite{guo2017region} for network details of Region Ensemble (REN). In our implementation, the final regressor in REN outputs $26$ dimensional $\delta \theta$.    
The ShapeCNN consists of two simpler CNNs similar to \cite{oberweger2015hands}; called $\alpha$-CNN and $\beta$-CNN. 
Each of them  has $3$ convolutional layers using kernels sizes $5$,$5$,$3$ respectively. First two convolution layers are followed by max pool layers. The pooling layers use strides of $4$ and $2$. The convolutional layers generate $8$ feature maps of size $12$ x $12$. Lastly, the two fully connected (FC) layers have $1024$ neurons each with dropout ratio of $0.3$. After the second FC layer, the final FC layers in $\alpha$-CNN and $\beta$-CNN output $6$ dimensional $\alpha$ and $7$ dimensional $\beta$ parameters respectively. All layers use the ReLu as activation function. 


\subsection{Hand Pose and Shape Layer (HPSL)}
\label{ssec:HPSL}
HPSL is a non-linear differentiable layer, embedded inside the deep network as shown in Figure \ref{fig:application}(a). The task of the layer is to produce 3D joint positions  $P \in \mathcal{R}^ {3 \textrm{x} J}$ and vertices of hand mesh $V \in \mathcal{R}^ {3 \textrm{x} \vartheta}$ given the pose parameters $\Theta$, hand bones scales $\alpha$ and shape parameters $\beta$. The layer function can be written as: 
\begin{equation} \label{eq:1}
 \textit{(P,V)} = \textbf{\textrm{HPSL}}(\Theta,\beta,\alpha)  
\end{equation}


We compute the respective gradients in the layer for back-propagation.
The Euclidean 3D joint location and 3D vertex location losses are given as:
\begin{equation} \label{eq:22}
L_J = \frac{1}{2}\norm{ \textit{P} - {\textit{P}_G}_T }^2 \;\;\;\;\; , \;\;\;\;\;
L_V = \frac{1}{2}\norm{\textit{V} - {\textit{V}_G}_T }^2
\end{equation}
Where ${L}_J$ and ${L}_V$ are the 3D joint and vertex losses respectively. ${\textit{P}_G}_T$ and ${\textit{V}_G}_T$ are  vectors of 3D ground truth joint positions and mesh vertices, respectively.
Various functions inside the layer are detailed as follows:  

\noindent
\textbf{Hand Skeleton Bone-lengths Adaptation}:
In order to adapt bone-lengths of hand skeleton during training over varying hand shapes in the dataset, \cite{malik2017simultaneous} propose various bone-length scaling strategies. Following the similar approach, we assign a separate scale parameter for bone-lengths in palm $\bf{s_p}$ and $5$ different scales for bones  as shown in Figure \ref{fig:application}(b). The HPSL acquires the scaling parameters $\alpha=[\bf{s_p}, \bf{s_1}, \bf{s_2}, \bf{s_3}, \bf{s_4}, \bf{s_5}]$ from the ShapeCNN during the training process.  

\noindent
\textbf{Morphable Hand Model Formulation}:
Given the shape parameters $\bf{\beta}$ learned by our ShapeCNN, we reconstruct the hand shape surface by implementing a morphable hand model inside our HPSL.
A morphable hand model $\Psi \in \mathcal{R}^ {3 \textrm{x} \vartheta}$  is a set of 3D vertices representing a particular hand shape. Any morphable hand model can be expressed as a linear combination of principle hand shape components, called morphable targets $\bf{b_t}$ \cite{lewis2014practice}. 
Our principle hand shape components are defined for \textit{Length}, \textit{Mass}, \textit{Size}, \textit{Palm Length}, \textit{Fingers Inter-distance}, \textit{Fingers Length} and \textit{Fingers Tip-Size}. They represent offsets from a neutral hand shape $\bf{b_0}$ similar to one shown in Figure \ref{fig:application}(b). 
Each learned shape parameter $\bf{\beta_t}$  defines the amount of contribution of a principle shape components $\bf{b_t}$ towards formulation of final hand morphable model. Hence, a hand morphable model $\Psi$ can be formulated using the following Equation:
\begin{equation} \label{eq:2}
{\Psi} (\bf{\beta}) = \bf{b_0} + \sum_{t=1}^{7} \bf{\beta_t}(\bf{b_t}-\bf{b_0})
\end{equation}


\noindent
\textbf{Forward Kinematics and Geometric Skinning}:
To estimate the 3D hand joints positions and surface vertices, we implement forward kinematics and geometric skinning functions inside our HPSL. As this layer is part of our deep network, it is essential to compute and back-propagate the gradients of these functions. The rest of this section addresses the definition of these functions and their gradients. 

The deformation of the hand skeleton from the reference pose $\theta_{init}$ to the current pose $\Theta$  can be obtained by transforming each joint $j_i$ along the kinematic chain by simple rigid transformations matrices. In our algorithm, these  matrices are updated based on bones scales $\alpha$ and the changes in pose parameters $\delta \theta$ which are learned by our ShapeCNN and PoseCNN, respectively. The kinematics equation of joint $j_i$ can be written as: 
\begin{equation} \label{eq:3}
\begin{split}
  j_i & = {\textrm{F}_j}_i(\Theta,\alpha) = {\textrm{M}_j}_i [0,0,0,1]^T  \\
  & = \big(\prod_{k \in {S_j}_i} [{\textrm {R}_\phi}_k(\theta_k)] \times [ {\textrm {T}_\phi}_k (\alpha \textit{B})]\big)[0,0,0,1]^T 
 \end{split}
\end{equation}
where ${\textrm{M}_j}_i$ represents the transformation matrix from the zero pose (i.e.  joint at position $[0,0,0,1]$) to the  current pose. ${S_j}_i$ is the set of joints along kinematic chain from $j_i$ to the root joint and $\phi_k$ is one of the rotation axes of joint $k$.

For animating the 3D hand mesh, we use linear blend skinning \cite{lewis2000pose} to deform the set of vertices $\vartheta$ according to underlying hand skeleton kinematic transformations. The skinning weights $\omega_i$, define the skeleton-to-skin bindings. Their values represent the influence of joints on their associated vertices. Normally, the weights of each vertex  are assumed to be convex (i.e. $\sum_{i=1}^{n} \omega_i = 1$) and $\omega_i >0$. The transformation of a vertex $\textrm{v}_{\varkappa} \in \Psi$ can be defined as:
\begin{equation} \label{eq:5}
\begin{split}
 {\textrm{v}_\varkappa}  & = \Upsilon_{\textrm{v}_\varkappa} (\Theta,\beta,\alpha) =\sum_{i \in {P_\textrm{v}}_\varkappa} \omega_i \bf{{C_j}_i} \textrm{v}_\varkappa (\beta) \\
        & =\sum_{i \in {P_\textrm{v}}_\varkappa} \omega_i \bf{{C_j}_i} ({b_0^{v_\varkappa}} + \sum_{t=1}^{7} {\beta_t}({b_t^{v_\varkappa}}-{b_0^{v_\varkappa}}))
 \end{split}
\end{equation}
where ${P_\textrm{v}}_\varkappa$ is the set of joints influencing the vertex $\textrm{v}_\varkappa$  and $\bf{{C_j}_i}$ is the transformation matrix of each joint $j_i$ from its reference pose $\theta_{init}$ to its actual position in the current animated posture. $\bf{{C_j}_i}$ can be represented as:
\begin{equation} \label{eq:4}
\bf{{C_j}_i} = {\textrm{M}_j}_i {{\textrm{M}_j}_i^*}^{-1}   
\end{equation}
where  ${{\textrm{M}_j}_i^*}^{-1}$ defines the inverse of reference pose transformation matrix.

\noindent
\textbf{Gradients computation}: 
For backward-pass in the \textbf{HPSL}, we compute gradients of the following equation with respect to the layer inputs:
\begin{equation} \label{eq:1a}
\textbf{\textrm{HPSL}}(\Theta,\beta,\alpha) = \textit{( $\textbf{\textrm{F}}(\Theta,\alpha)$ ,$\Upsilon(\Theta,\beta,\alpha)$ )}. 
\end{equation}
Each vertex $\textrm{v}_{\varkappa} = \textrm{HPSL}_{\textrm{v}_\varkappa}(\Theta,\beta,\alpha)$ in the reconstructed hand morphable model $\Psi$ is deformed using Equation \ref{eq:5}. 
Hence, its gradients with respect to a shape parameter $\beta_t$ can be computed as: 
\begin{align*} 
\frac{\partial (\textrm{HPSL}_{\textrm{v}_\varkappa})}{\partial \beta_t} &= 
\sum_{i} \omega_i \bf{{C_j}_i} (b_t^{v_\varkappa}-b_0^{v_\varkappa})&  &\text{for}\ t = 1,2,\ldots,7
\end{align*}
According to Equation \ref{eq:1a}, bones scales influence the joints positions and vertices positions. Hence, the resultant gradient with respect to a hand scale parameter $\alpha_s$, can be calculated as:
\begin{align*} 
\frac{\partial (\textbf{\textrm{HPSL}})}{\partial \alpha_s} &= 
\frac{\partial \textbf{\textrm{F}}}{\partial \alpha_s} + \frac{\partial \Upsilon}{\partial \alpha_s}&  &\text{for}\ s = 1,2,\ldots,6
\end{align*}
To compute the partial derivative of $\textbf{\textrm{F}}$ with respect to $\alpha_s$, we need to derivate each joint with respect to its associated scale parameter. The gradient of a joint with respect to $\alpha_s$, can be computed by replacing the scaled translational matrix containing $\alpha_s$ by its derivative and keep all other matrices same; see Equation $2$ in supplementary document. In a similar way, the gradient of a vertex $\textrm{v}_\varkappa$ with respect to $\alpha_s$ can be computed by:
\begin{align*} 
\frac{\partial \Upsilon_{\textrm{v}_\varkappa}}{\partial \alpha_s} & = \sum_{i} \omega_i  \frac{\partial \bf{{C_j}_i}}{\partial \alpha_s} \textrm{v}_\varkappa \\
& = \sum_{i} \omega_i [ {\textrm{M}_j}_i ({{\textrm{M}_j}_i^*}^{-1})' + ({\textrm{M}_j}_i)' {{\textrm{M}_j}_i^*}^{-1}] \textrm{v}_\varkappa & 
\end{align*}
Likewise, for the pose parameters $\Theta$, we compute the following equation: 
\begin{align*} 
\frac{\partial (\textbf{\textrm{HPSL}})}{\partial \theta_p} &= 
\frac{\partial \textbf{\textrm{F}}}{\partial \theta_p} + \frac{\partial \Upsilon}{\partial \theta_p}&  &\text{for}\ p = 1,2,\ldots,26
\end{align*}
Accordingly, the derivative of a joint with respect to a pose parameter $\theta_p$, is simply to replace the rotation matrix of $\theta_p$ by its derivation; see Equation $5$ in supplementary document. And, the derivative of a vertex $\textrm{v}_\varkappa$ with respect to $\theta_p$ is computed by:     
\begin{align*} 
\frac{\partial \Upsilon_{\textrm{v}_\varkappa}}{\partial \theta_p} & = \sum_{i} \omega_i  \frac{\partial \bf{{C_j}_i}}{\partial \theta_p} \textrm{v}_\varkappa \\
& = \sum_{i} \omega_i  [({\textrm{M}_j}_i)' {{\textrm{M}_j}_i^*}^{-1}  ] \textrm{v}_\varkappa&  &\text{for}\ p = 1,2,\ldots,26
\end{align*}
More details about the gradients computation can be found in the supplementary document.


\section{Synthetic Dataset}
\label{sec:SYN}

There are two main objectives of creating our synthetic dataset. First is to jointly recover full hand shape surface and pose provided that there is no ground truth hand surface information available in public benchmarks; see Section \ref{ssec:Experiments_a}. Second objective is to provide a training data with sufficient variation in hand shapes and poses such that a CNN model can be pre-trained to improve the recognition rates on real benchmarks; see Section \ref{ssec:Experiments_b}. This problem is different from generating very realistic hand-shape, where a real-world statistical hand model~\cite{MANO:SIGGRAPHASIA:2017} can be applied. 
However, the variation in shape is more challenging for real-world databases
e.g. BigHand2.2M \cite{yuan2017bighand2} database was captured from only $10$ users, and the MANO ~\cite{MANO:SIGGRAPHASIA:2017} database was built from the contribution of $31$ users.
Instead, we generate a bigger hand shape variation which may not be present in a given cohort of human users.

Our SynHand5M dataset offers $4.5M$ train and $500K$ test images; see Figure \ref{fig:datasetVC}(a) for SynHand5M components.
SynHand5M uses the hand model generated by ManuelBastionLAB \cite{ManuelBastionLAB} which is a procedural full-body generator distributed as add-on of the Blender \cite{blender} 3D authoring software.  
Our virtual camera simulates a Creative Senz3D Interactive Gesture Camera \cite{camera_Senz3D}. It renders images of resolution 320x240 using diagonal field of view of 74 degrees.
In the default position, the hand palm faces the camera orthogonally and the fingers point up.
We procedurally modulate many parameters controlling the hand and generate images by rendering the view from the virtual camera. The parameters characterizing the hand model belong to three categories: hand shape, pose and view point. 

Without constraints the hand generator can easily lead to impossible hand shapes. So, in order to define realistic range limits for modulating hand shapes, we relied on the DINED \cite{molenbroek04dined} anthropometric database. DINED is a repository collecting the results of several anthropometric databases, including the CAESAR surface anthropometry survey \cite{robinette_caesar_1999}. 
We manually tuned the ranges of the $7$ hand shape parameters (see Section \ref{ssec:HPSL}) in order to cover 99\% of the measured  population in this dataset; see supplementary document for more details. 


To modulate the hand pose, we manipulate the $26$ DoFs of our hand model; see Figure \ref{fig:application}(b). For each finger, rotations are applied to flexion of all phalanges plus the abduction of the proximal phalanx. Additionally, in order to increase the realism of the closed fist configuration, the roll of middle, ring, and pinky fingers is derived from the abduction angle of the same phalanx.
The rotation limits are set to bring the hand from a closed fist to an over-extended aperture, respecting anatomical constraints and avoiding the fingers to enter the palm.

The hand can rotate about three DoFs to generate different view points: roll around its longitudinal axis (i.e. along the fingers), rotate around the palm orthogonal axis (i.e. rolling in front of the camera), and rotate around its transversal axis (i.e. flexion/extension of the wrist).
\section{Experiments and Results}
\label{sec:Experiments}

In this section, we provide the implementation details, quantitative and qualitative evaluations of the proposed algorithm and the proposed dataset.  
We use three evaluation metrics; mean 3D joint location error (JLE), 3D vertex location error (VLE) and percentage of images within certain thresholds in $mm$.

Recent CNN-based discriminative methods such as \cite{ge2017robust, wang2018region, moon2017v2v,rad2017feature} outperform CNN-based hybrid methods; see Section \ref{sec:related}. However, due to direct joints regression, discriminative methods neither explicitly account for the hand shapes nor consider kinematics constraints \cite{zhou2016model, malik2017simultaneous}. Moreover, in contrast to hybrid methods, discriminative methods generalize poorly to unseen hand shapes; see \cite{yuan2018depth}. 
Our proposed hybrid method does not exceed in accuracy over recent discriminative works but, it does not suffer from such limitations. Therefore, it is not fair to compare with these methods.  
However, we compare with the state-of-the-art hybrid methods and show improved performance.   
Notably, we propose the first algorithm that jointly regresses hand pose, bone-lengths and shape surface in a single network.

\begin{figure}
\centering
\includegraphics[width=4.1cm,height=3.4cm]{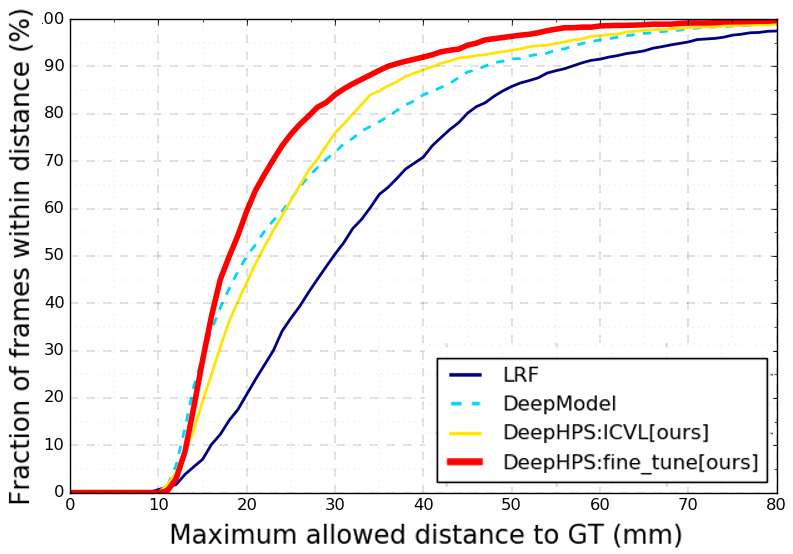}
\includegraphics[width=4.1cm,height=3.4cm]{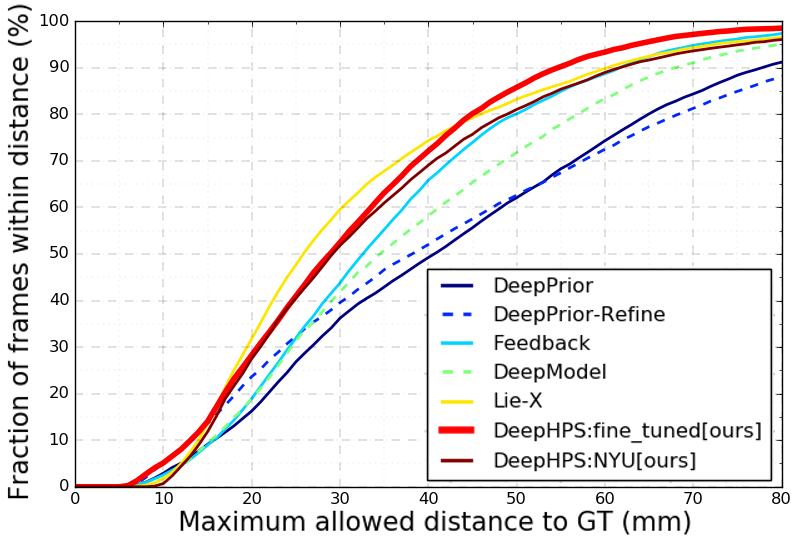}
     (a) ICVL \;\;\;\;\;\;\;\;\;\;\;\;\;\;\;\;\;\;\;\;\;\;\;\;\;\;\;\;\; (b) NYU 
\caption{\textbf{Quantitative evaluation}. (a) show the results of our algorithm (DeepHPS) on ICVL test set, when trained on ICVL and fine-tuned on ICVL. (b) is the same but with NYU. To fine-tune, we pretrain DeepHPS on our SynHand5M. Our results on ICVL show improved accuracy over the state-of-the-art hybrid methods (e.g. LRF\cite{tang2014latent} and DeepModel\cite{zhou2016model}). On NYU, the results are better than the state-of-the-art hybrid methods (e.g. DeepPrior\cite{oberweger2015hands}, DeepPrior-Refine\cite{oberweger2015hands}, Feedback\cite{oberweger2015training}, DeepModel\cite{zhou2016model} and Lie-X\cite{xu2017lie}). The curves show the number of frames in error within certain thresholds.}
\label{fig:cross-benchmark}
\vspace{-5mm}
\end{figure}

\subsection{Implementation Details}
For training, we pre-process the raw depth data for standardization and depth invariance. We start by computing the centroid of the hand region in the depth image. The obtained 3D hand center location (i.e. palm center) is used to crop the depth frame. The camera intrinsics (i.e. focal length) and a bounding box of size $150$, are used during the crop. The pre-processed depth image is of size $96$ x $96$ and in depth range of $[$$-1$, $1$$]$. The annotations in camera coordinates are simply normalized by the bounding box size and clipped in range $[$$-1$, $1$$]$.  

We use Caffe \cite{jia2014caffe} which is an open-source training framework for deep networks. The complete pipeline is trained end-to-end until convergence. The learning rate was set to $0.00001$ with 0.9 SGD momentum. A batch size of $256$ was used during the training. The framework is executed on a desktop equipped with Nvidia Geforce GTX $1080$ Ti GPU with 16GB RAM. One forward pass takes $3.7ms$ to generate 3D hand joint positions and shape surface. For simplicity, we name our method as DeepHPS.       


\begin{figure}
\begin{center}
    \setlength\fboxsep{0.6pt}
    \setlength\fboxrule{0.6pt}
      \includegraphics[width=0.95\linewidth]{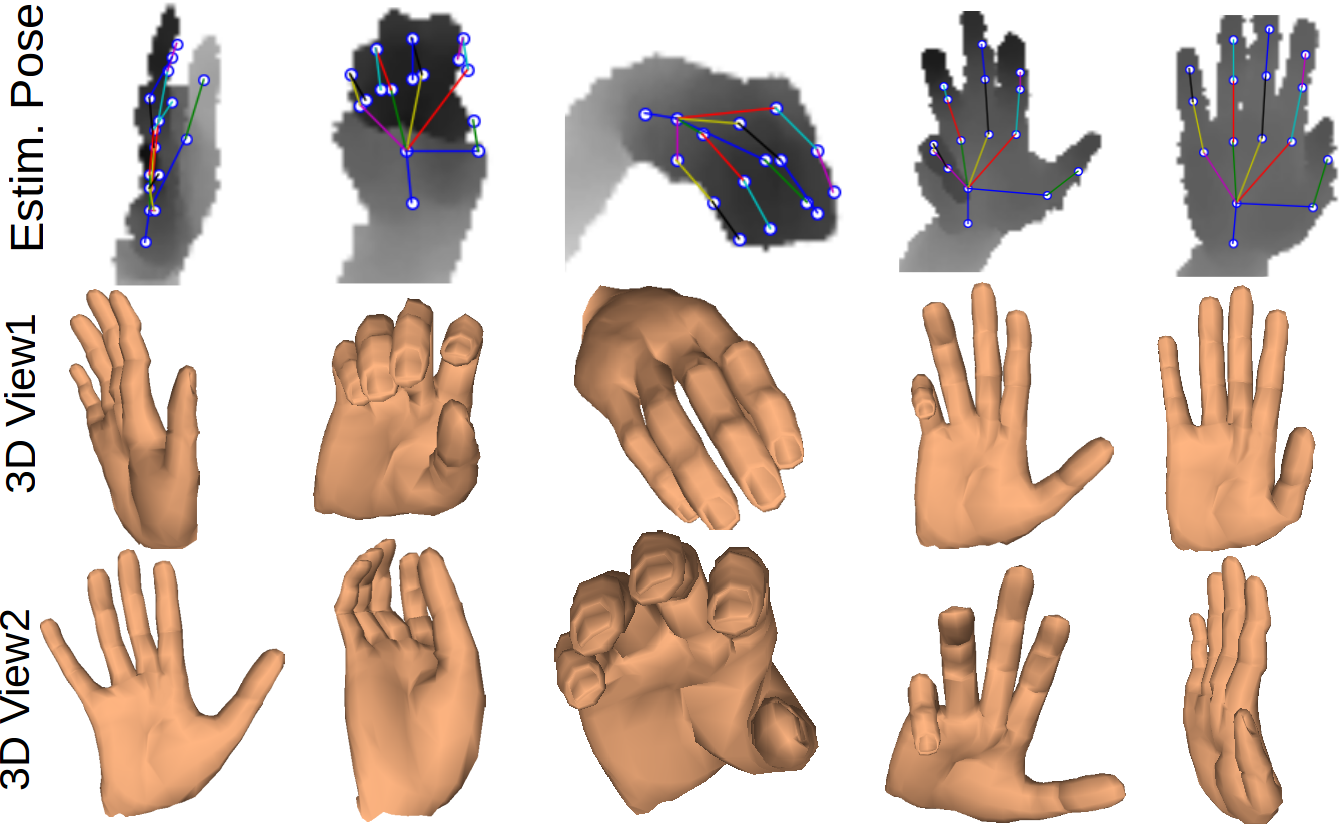}
\end{center}
    \vspace{-5mm}
\caption{ {\bf{Real hand pose and shape recovery}}: More results on hand pose and surface reconstruction of NYU\cite{tompson2014real} images. Despite of unavailability of ground truth hand mesh vertices, our algorithm produces plausible hand shape. 
   }
    \vspace{-4.5mm}
\label{fig:figure_title_1}
\end{figure}

\subsection{Algorithm Evaluation}
\label{ssec:Experiments_a}
In this subsection, we evaluate our complete pipeline using the SynHand5M. Moreover, we devise a joint training strategy for both real and synthetic datasets to show qualitative hand surface reconstruction of real images. 

\begin{figure*}[t]
\begin{center}
	\setlength\fboxsep{0.6pt}
	\setlength\fboxrule{0.6pt} 
      \includegraphics[width=0.86\linewidth]{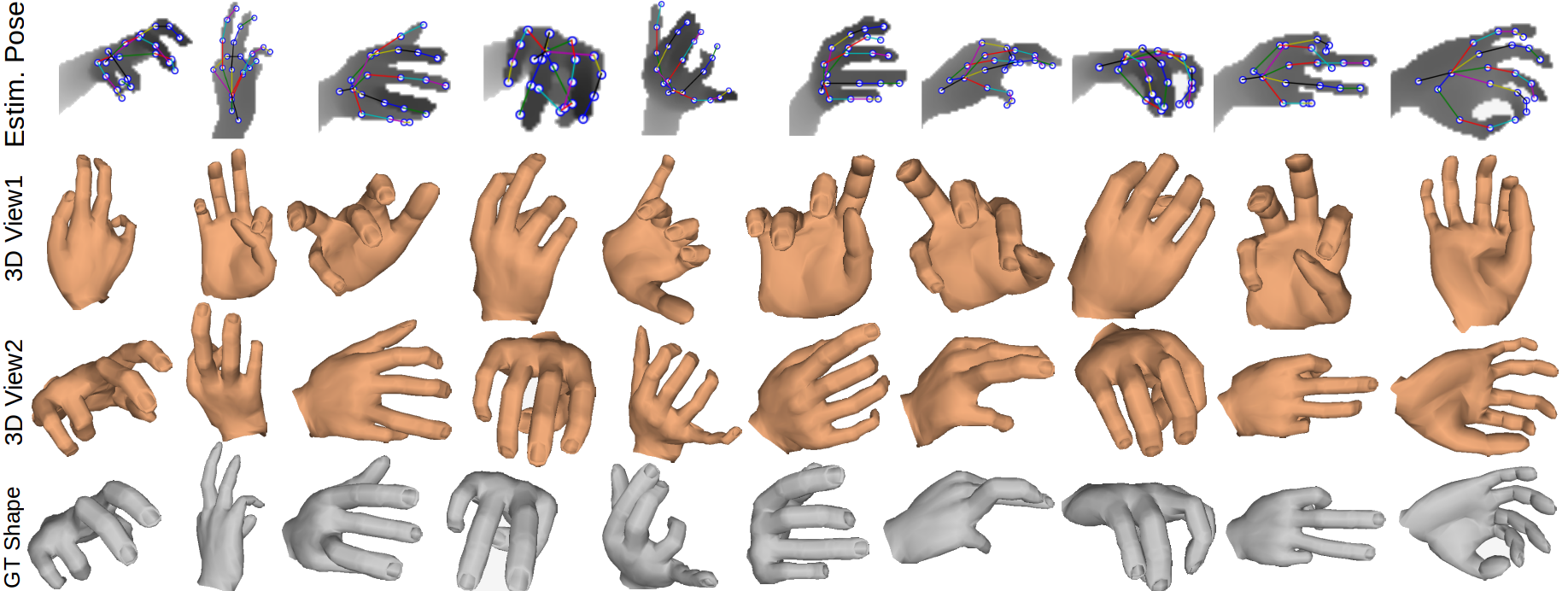}
      \includegraphics[width=0.86\linewidth]{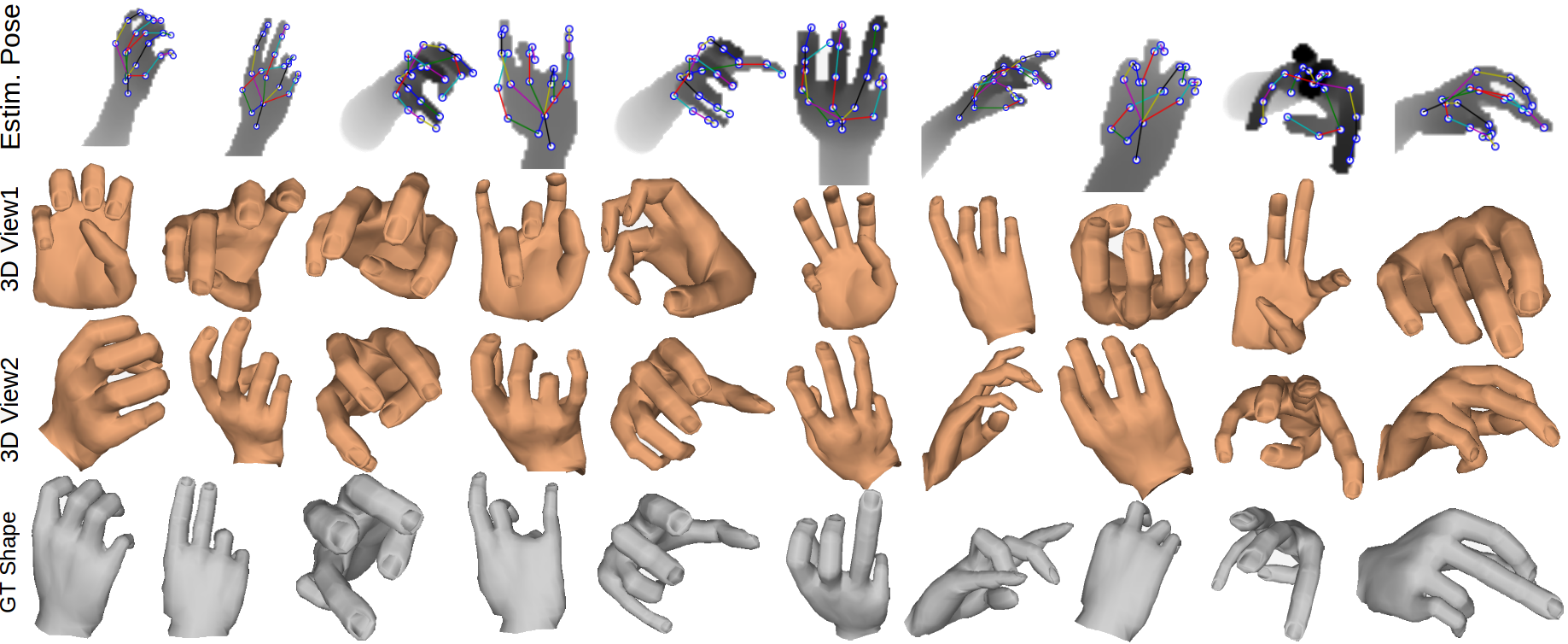}
\end{center}
    \vspace{-5mm}
   \caption{ \textbf{Synthetic hand pose and shape recovery}: We show example estimated hand poses overlaid with the preprocessed depth images from our SynHand5M. We show the reconstructed surface from two different views (yellow) and  the ground truth surface (gray). \textbf{\textit{3D View2}} is similar to the ground truth view. Our algorithm infers correct 3D pose and shape even in very challenging condition, like  occlusion of several fingers and large variation in view points. 
   }
\label{fig:Synthetic_results}
\vspace{-5mm}
\end{figure*}

\noindent
\textbf{Evaluation on the synthetic dataset}:
The complete pipeline is trained end-to-end using SynHand5M for pose and shape recovery. For fair comparison, we train the state-of-the-art model based learning methods \cite{zhou2016model, malik2017simultaneous} on SynHand5M. \cite{malik2017simultaneous} works for varying hand shapes in contrast to the closely related method \cite{zhou2016model}. The quantitative results are shown in Table \ref{tab:cross-benchmark}. Our method clearly exceeds in accuracy over the compared method and additionally reconstructs full hand surface. 
The qualitative results are shown in Figure \ref{fig:Synthetic_results}. 
The estimated $22$ joint positions are overlaid on the depth images while the reconstructed hand surface is shown using two different views named as \textbf{\textit{3D View1}} and \textbf{\textit{3D View2}}. For better visualization, view2 is similar to ground truth view. 
The results demonstrate that our DeepHPS model infers correct hand shape surface even in cases of occlusion of several fingers and large variation in view points.
    
\noindent
\textbf{Evaluation on the NYU real dataset}: In order to jointly train our whole pipeline on both real and synthetic data, we found $16$ closely matching common joint positions in SynHand5M and the NYU dataset. These common joints are different from the $14$ joints used for the public comparisons \cite{tompson2014real}. The loss equation is; 
\vspace{-2mm}  
\begin{equation} \label{eq:20}
{L} = {L}_J + \mathbbm{1}{L}_V  
\vspace{-2mm} 
\end{equation}

where $\mathbbm{1}$ is an indicator function which specifies whether the ground truth for mesh vertices is available or not. In our setup, it is 1 for synthetic images and 0 for real images. For real images, backpropagation from surface reconstruction part is disabled. 

The qualitative pose and surface shape results on sample NYU real images are shown in Figure \ref{fig:figure_title} and \ref{fig:figure_title_1}. Despite of the missing ground truth surface information and presence of high camera noise in NYU images, the resulting hand surface is plausible and the algorithm performs well in case of missing depth information and occluded hand parts.

\begin{table}[htb]
\begin{center}
\begin{tabular}{|l|c|c|}
\hline
Method \textbackslash Error(mm)  	& 3D Joint Loc.  & 3D Vertex Loc. \\
\hline\hline
DeepModel \cite{zhou2016model} & 11.36 & -- \\
\hline
HandScales \cite{malik2017simultaneous} & 9.67 & -- \\
\hline
DeepHPS [\textbf{Ours}] & {\bf 6.3} & {\bf 11.8}\\
\hline
\end{tabular}
\end{center}
\vspace{-5mm}
\caption{\textbf{Quantitative Evaluation on SynHand5M}: We show the 3D joint and vertex locations errors(mm). Our method additionally outputs mesh vertices and outperforms model based learning methods \cite{zhou2016model,malik2017simultaneous}.}
\label{tab:cross-benchmark}
\end{table}

\begin{table}[htb]
\begin{center}
\begin{tabular}{|l|c|}
\hline
Methods & 3D Joint Location Error \\
\hline\hline
DeepPrior \cite{oberweger2015hands} & 20.75mm \\
DeepPrior-Refine \cite{oberweger2015hands} & 19.72mm \\
Crossing Nets \cite{wan2017crossing} & 15.5mm \\
Feedback \cite{oberweger2015training} & 15.9mm \\
DeepModel \cite{zhou2016model} & 17.0mm \\
Lie-X \cite{xu2017lie} & 14.5mm \\
\detokenize{DeepHPS:NYU} [Ours] & 15.8mm \\
\detokenize{DeepHPS:fine-tuned} [\textbf{Ours}] & {\bf 14.2mm} \\
\hline
\end{tabular}
\end{center}
\vspace{-5mm}
\caption{\textbf{Quantitative comparison on NYU} \cite{tompson2014real}: Our fine-tuned DeepHPS model on the NYU dataset shows the state-of-the-art performance among hybrid methods.}
\label{tab:NYU}
\end{table}

\begin{table}[htb]
\begin{center}
\begin{tabular}{|l|c|}
\hline
Methods & 3D Joint Location Error \\
\hline\hline
LRF \cite{tang2014latent} & 12.57mm \\
DeepModel \cite{zhou2016model} & 11.56mm \\
Crossing Nets \cite{wan2017crossing} & 10.2mm \\
\detokenize{DeepHPS:ICVL} [Ours] & 10.5mm \\
\detokenize{DeepHPS:fine-tuned} [\textbf{Ours}] & {\bf 9.1mm} \\
\hline
\end{tabular}
\end{center}
\vspace{-5mm}
\caption{\textbf{Quantitative comparison on ICVL} \cite{tang2014latent}: The DeepHPS model fine-tuned on the ICVL dataset outperforms the state-of-the-art hybrid methods.}
\label{tab:ICVL}
\vspace{-5mm}
\end{table}


\subsection{Comparison on Public Benchmarks}
\label{ssec:Experiments_b}
The public benchmarks do not provide ground truth hand mesh files. Therefore,
we provide quantitative results for pose inference on two of the real hand pose datasets (i.e. NYU and ICVL).
For comparisons, NYU dataset use $14$ joint positions \cite{tompson2014real} whereas ICVL dataset \cite{tang2014latent} use $16$ joint positions. 


Our DeepHPS algorithm is trained on NYU and ICVL individually, called DeepHPS:NYU and DeepHPS:ICVL models. Then, we fine-tune the pre-trained DeepHPS (on SynHand5M) with the NYU and ICVL, we call DeepHPS:fine-tuned models. The 3D joint location errors of the trained models are calculated on $8252$ NYU and $1596$ ICVL test images respectively. The quantitative results are shown in Figure \ref{fig:cross-benchmark} and Tables \ref{tab:NYU} and \ref{tab:ICVL}. DeepHPS:fine-tuned models achieve an error improvement of $13.3\%$ and $10.12\%$ over DeepHPS:ICVL and DeepHPS:NYU models respectively. 

On the ICVL and NYU datasets, we achieve improvement in the joint location accuracy over the state-of-the-art hybrid methods.
 
\begin{figure}[t]
\begin{center}
\includegraphics[width=0.2\linewidth]{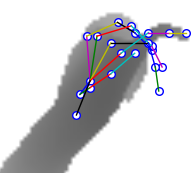} \;\;
\includegraphics[width=0.2\linewidth]{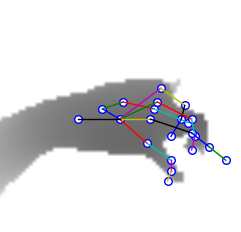} \;\;\;\;\;\;\;\;\;\;
\includegraphics[width=0.09\linewidth]{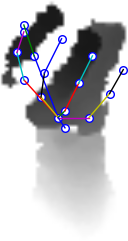} \;\;\;\;
\includegraphics[width=0.12\linewidth]{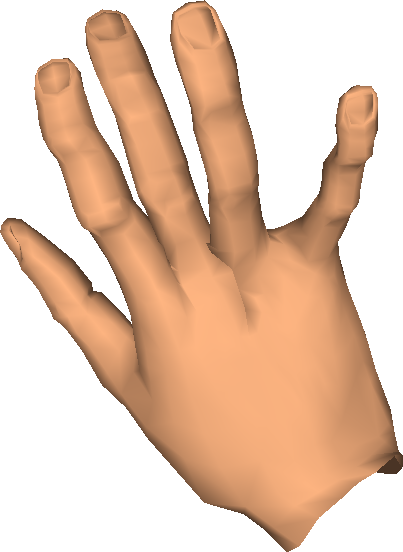}
\;\;\;\;\;\;\;\;\;\; (a)  \;\;\;\;\;\;\;\;\;\;\;\;\;\;\;\;\;\;\;\;\;\;\;\;\;\;\;\;\;\;\;\;\;\;\;\;(b) \;\;
\end{center}
    \vspace{-6mm}
   \caption{ {\bf{Failure case}}:(a) incorrect pose due to highly occluded hand parts. (b) incorrect pose and shape due to significant missing depth information.  
   }
\label{fig:failure_case}
\vspace{-5mm}
\end{figure}

\noindent
\textbf{Failure case}:   
Our framework works well in case of missing depth information and occlusions.
However, under severe occlusions and a lot of missing depth information, it may fail to detect the correct pose and shape; see Figure \ref{fig:failure_case}. 

\section{Conclusion and Future Work}
In this work, we demonstrate the simultaneous recovery of hand pose and shape surface from a single depth image. For training, we synthetically generate a large scale dataset with accurate joint positions, segmentation masks and hand meshes of depth images. Our dataset will be a valuable addition for training and testing CNN-based models for 3D hand pose and shape analysis. Furthermore, it improves the recognition rate of CNN models on hand pose datasets. 
In our algorithm, intermediate parametric representations are estimated from a CNN architecture. Then, a novel hand pose and shape layer is embedded inside the deep network to produce 3D hand joint positions and shape surface. Experiments show improved accuracy over the state-of-the-art hybrid methods. Furthermore, we demonstrate plausible results for the recovery of hand shape surface on real images. 
Improving the performance of CNN-based hybrid methods is a potential research direction. These methods bear a lot of potential due to their inherent stability and scalability. In future, we wish to extend our dataset with wider view points coverage, object interactions and RGB images. Another aspect for future work is predicting fine-scale 3D surface detail on the hand, where real-world statistical hand models~\cite{MANO:SIGGRAPHASIA:2017} possibly give better priors.
\vspace{-2ex}

\section*{Acknowledgements}
\vspace{-1ex}
This work was partially funded by NUST, Pakistan, the Federal Ministry of Education and Research
of the Federal Republic of Germany as part of the research projects DYNAMICS (Grant number 01IW15003) and VIDETE (Grant number 01IW18002).

{\small
\bibliographystyle{ieee}
\bibliography{egbib}
}

\end{document}